\documentclass{article}

\usepackage[preprint]{neurips_2026}
\usepackage{graphicx}
\usepackage[utf8]{inputenc}
\usepackage[T1]{fontenc}
\usepackage{hyperref}
\usepackage{url}
\usepackage{booktabs}
\usepackage{amsfonts}
\usepackage{amsmath}
\usepackage{nicefrac}
\usepackage{microtype}
\usepackage{xcolor}
\usepackage{multirow}

\title{Playing games with knowledge: AI-Induced delusions need game theoretic interventions}

\author{%
  Will Beaumaster \quad Paul Schrater\thanks{Code available at \url{https://github.com/willbeau1234/Epistemic-mediator}}\\
  University of Minnesota
}

\begin{document}

\maketitle

\begin{abstract}
Recent literature has identified a fundamental flaw in conversational AI: sycophantic chatbots induce delusional belief spirals even in rational agents. We argue that this ``epistemic entrenchment'' is not merely a model failure, but a systemic consequence of the paradigm shift from static search to strategic, repeated-play communication. We formalize this interaction as a Crawford-Sobel cheap talk game, where costless user signals induce a pooling equilibrium: sycophantic agents, optimized for user satisfaction, provide identical reinforcement to both exploratory ``Growth-seekers'' ($\theta_G$) and confirmatory ``Validation-seekers'' ($\theta_V$). Under repeated play, this identification failure creates a coordination trap---analogous to a Prisoner's Dilemma---where locally rational feedback loops drive users toward pathologically certain false beliefs. 
We propose the Epistemic Mediator, an inference-time mechanism design intervention that breaks this pooling equilibrium by introducing epistemic friction. This friction serves as a costly signal, forcing type revelation based on the users' asymmetric cognitive costs for processing resistance. Our primary contribution is Belief Versioning, a git-inspired epistemic memory system that commits belief states at healthy moments and executes a ``rollback'' when validation-seeking resistance is detected. In simulation, this intervention achieves a separating equilibrium where heterogeneous agents exhibit a $48\times$ differential in spiral rates ($0.8\%$ vs. $38.7\%$). Belief Versioning reduces spiral rates from $53.6\%$ to $9.0\%$ while passing a ``Learning Preservation Criterion'' (mean belief $\bar{P} = 0.32$). We validate these findings in GPT-4o, where Belief Versioning reduces spiral rates from $100\%$ to $16.5\%$. Our results demonstrate that epistemic safety in AI is fundamentally a problem of strategic information environment design rather than simple model alignment.
\end{abstract}

\section{Introduction}

Users approaching conversational AI systems present a fundamental identification
problem: truth-seekers and validation-seekers produce identical input signals
despite having opposite epistemic motivations. Consider two users who each tell
a chatbot ``I think my neighbor is watching me.'' The first is genuinely
investigating an uncertain belief and wants to reason carefully about the
evidence. The second has already concluded their neighbor is a spy and is seeking
confirmation. These two types---which we term Growth-seekers ($\theta_G$) and
Validation-seekers ($\theta_V$)---send identical signals to the AI system, yet
require fundamentally different responses.

Current large language models cannot distinguish between these types. Trained
via reinforcement learning from human feedback \citep{ouyang2022training},
modern LLMs exhibit a well-documented tendency toward sycophancy: excessively
agreeing with user beliefs regardless of their validity
\citep{sharma2023understanding, wei2023simple}. For truth-seekers, this is
mildly suboptimal. For validation-seekers, this is pathological. Recent work
has demonstrated that sycophantic feedback can induce delusional belief spirals
even in agents that update their beliefs rationally \citep{chandra2026sycophantic}.
As LLMs are deployed at scale in therapeutic, advisory, and companionship
settings, the inability to distinguish these types creates a systematic mechanism
for reinforcing pathological beliefs to the point of high false certainty.

We formalize this as a \textit{cheap talk} problem from Crawford-Sobel game
theory \citep{crawford1982strategic}. When signals are costless, senders with
different types send identical messages because there is no incentive to reveal
type. A sycophantic AI operating in this pooling equilibrium reinforces both
types identically, producing what we term \textit{epistemic entrenchment}:
the progressive collapse of a user's belief distribution toward pathological
certainty under sustained sycophantic feedback. Existing approaches to
sycophancy mitigation fall into two categories: training-time interventions
that require model access and retraining \citep{wei2023simple}, and measurement
frameworks that diagnose the problem without addressing it \citep{atwell2025basil}.
Neither operates at inference time without modifying the underlying model, and
neither addresses the user belief dynamics that produce pathological outcomes.

We propose the \textbf{Epistemic Auditor}, an inference-time architecture
requiring no model retraining, that breaks this pooling equilibrium by
introducing epistemic friction when real-time monitoring detects the dynamical
signature of spiral onset. The key insight is that delusional spirals have a
characteristic signal in belief dynamics: entropy decays while confidence
accelerates. By monitoring these quantities continuously, the Auditor detects
spiral onset and injects calibrated friction, forcing a costly signal that
separates $\theta_G$ from $\theta_V$ users through their differential response
to epistemic cost.

Our primary contribution is \textbf{Belief Versioning}: a git-inspired epistemic
memory system that commits belief states at epistemically healthy moments and
rolls back when validation-seeking resistance is detected. Unlike continuous
friction approaches that suppress all belief movement, Belief Versioning preserves
genuine epistemic updating while interrupting pathological entrenchment. We also
identify a critical failure mode in naive predictive controllers: achieving 0\%
spiral rates by driving mean belief to maximum uncertainty ($P \approx 0.50$)
is a trivial solution that destroys the learning the system is meant to protect.

Our contributions are as follows:
\begin{itemize}
    \item \textbf{Formal model:} We formalize sycophancy-induced belief
    entrenchment as a dynamical system over Bayesian agent beliefs, grounded
    in Crawford-Sobel cheap talk theory, and characterize the pooling
    equilibrium failure that enables delusional spirals
    (Section~\ref{sec:model}).

    \item \textbf{Detection finding:} Through threshold ablation across 16
    parameter combinations, we show that entropy decay $\Delta\mathcal{H}$
    is the dominant detectable signature of spiral onset, with entrenchment
    velocity $V_e$ providing no additional detection power in normal operating
    ranges (Section~\ref{sec:experiments}).

    \item \textbf{Reactive intervention:} The Reactive Auditor reduces
    catastrophic belief entrenchment from 53.6\% to 16.6\%
    ($z = 17.334$, $p \approx 0$, 95\% CI non-overlapping) with a mean of
    4.1 interventions per 50-turn conversation, establishing a strong
    baseline (Section~\ref{sec:experiments}).

    \item \textbf{Belief Versioning (primary contribution):} Our git-inspired
    epistemic memory system reduces spiral rates to 9.0\% (83\% reduction)
    while preserving genuine belief updating (mean final belief $\bar{P}=0.32$
    vs.\ $\bar{P}=0.50$ for suppression-based approaches), generalizes
    out-of-distribution across sycophancy levels $p_\chi \in \{60,70,80,90\}$
    and longer time horizons (Section~\ref{sec:experiments}).

    \item \textbf{Type separation:} Heterogeneous agent simulations reveal a
    48$\times$ differential in spiral rates between $\theta_G$ and $\theta_V$
    users (0.8\% vs.\ 38.7\%), providing empirical evidence for the theorized
    separating equilibrium and validating the game-theoretic framing
    (Section~\ref{sec:experiments}).

    \item \textbf{Failure mode identification:} We demonstrate that continuous
    friction controllers achieving 0\% spiral rates do so by suppressing all
    belief movement (mean belief $\bar{P} \approx 0.50$), constituting a
    trivial solution that fails as an epistemic intervention. We provide a
    diagnostic criterion distinguishing genuine spiral suppression from
    learning suppression (Section~\ref{sec:experiments}).

    \item \textbf{LLM validation:} We validate simulation findings in GPT-4o
    ($n=200$) under high-sycophancy deployment configurations, demonstrating
    that Belief Versioning reduces spiral rates from 100\% to 16.5\%
    while outperforming the Reactive Auditor by 30.5 percentage points
    ($z=6.552$, $p=5.68\times10^{-11}$, large effect), establishing
    proof of concept for inference-time epistemic auditing in production
    systems without model retraining (Section~\ref{sec:experiments}).
\end{itemize}

\section{Related Work}
\label{sec:related}

Sycophancy in large language models has been documented across a range of
settings. \citet{sharma2023understanding} demonstrate that RLHF-trained models
systematically agree with user assertions even when those assertions are
factually incorrect, while \citet{wei2023simple} show that sycophantic behavior
persists across model scales and is reinforced by standard training objectives.
\citet{ouyang2022training} provide the mechanistic account: RLHF optimizes for
human approval, and human raters reliably prefer responses that validate their
existing beliefs. The consequence is a systematic misalignment between user
preference and epistemic benefit.

Existing mitigation approaches fall into two categories that each fail to
address the user belief dynamics we study. Training-time interventions
\citep{wei2023simple} require model access and retraining, making them
inapplicable to black-box API deployments and unable to adapt to individual
user belief trajectories. Measurement frameworks such as \citet{atwell2025basil}
provide diagnostic tools for identifying sycophancy but offer no intervention
mechanism. Neither category operates at inference time on the belief state
dynamics that produce pathological entrenchment.

The most directly related work is \citet{chandra2026sycophantic}, who prove
that sycophantic feedback induces delusional belief spirals even in agents that
update their beliefs rationally via Bayes' rule. Their result establishes that
the problem is not a failure of rationality but a failure of the information
environment: a sycophantic bot provides systematically biased evidence that
drives rational posteriors toward false certainty. We adopt their simulation
framework and extend it with intervention architectures, heterogeneous agent
models, and LLM validation.

Our game-theoretic framing draws on Crawford-Sobel cheap talk theory
\citep{crawford1982strategic}, which characterizes the conditions under which
costless signals fail to transmit information in equilibrium. To our knowledge
this is the first application of cheap talk pooling equilibrium analysis to the
LLM-user epistemic interaction, providing a formal account of why sycophantic
AI systems fail to distinguish truth-seekers from validation-seekers and
motivating the friction-based intervention we propose.

\section{Formal Model}
\label{sec:model}

\subsection{State Space}

We model a conversational AI interaction as a discrete-time dynamical system.
The world exists in one of two states:
\begin{equation}
    H \in \{H_0, H_1\}
\end{equation}
where $H_0$ denotes the null hypothesis (e.g., ``neighbor is not a spy'') and
$H_1$ denotes the alternative (e.g., ``neighbor is a spy''). The agent maintains
a joint belief distribution over hypothesis and bot sycophancy level:
\begin{equation}
    P_t(H, \chi) \quad \text{where } \chi \in [0, 1]
\end{equation}
The marginal belief at time $t$ is:
\begin{equation}
    P_t(H=1) = \sum_{\chi} P_t(H=1, \chi)
\end{equation}
User type is hidden and drawn at the start of each interaction:
\begin{equation}
    \theta \sim \text{Bernoulli}(p_V), \quad \theta \in \{\theta_G, \theta_V\}
\end{equation}
where $\theta_G$ denotes a Growth-seeker and $\theta_V$ a Validation-seeker.

\subsection{Likelihood Model}

The world generates binary evidence each turn. Each observation $d_i$ is drawn
from:
\begin{equation}
    P(d_i \mid H=0) = \text{Ber}(0.4), \qquad P(d_i \mid H=1) = \text{Ber}(0.6)
\end{equation}
The joint likelihood over $N=2$ independent observations is:
\begin{equation}
    P(d \mid H) = \prod_{i=1}^{N} \text{Ber}\!\left(d_i,\; \phi_{H,i}\right)
\end{equation}
The signal is deliberately weak: 0.6 vs.\ 0.4 means each observation provides
only mild evidence, modeling real-world ambiguity in which sycophancy can
overwhelm weak signals.

\subsection{Sycophantic Bot and the Pooling Equilibrium}

The bot's character is drawn each turn:
\begin{equation}
    \chi \sim \text{Bernoulli}\!\left(\frac{p_\chi}{100}\right)
\end{equation}
A \textsc{Fair} bot selects observation $o^*$ to maximize information gain. A
\textsc{Syco} bot selects $o^*$ to maximize the probability the human retains
their current hypothesis:
\begin{equation}
    o^*_{\textsc{Syco}} = \arg\max_{o} \;\Pr[\text{human retains }
    h_{\text{human}} \mid o]
\end{equation}
This is the \textit{cheap talk} pooling equilibrium failure
\citep{crawford1982strategic}. When signals are costless, the sycophantic bot's
incentives are misaligned with truth, so its signal carries no information about
reality in equilibrium. Both $\theta_G$ and $\theta_V$ users receive identical
responses, making type identification impossible without a costly signal.

\subsection{Bayesian Belief Update}

After observing bot output $(o, v)$, the agent updates their joint belief via
Bayes' rule:
\begin{equation}
    P_{t+1}(H, \chi) = \frac{P(o, v \mid H, \chi, d) \cdot P_t(H, \chi)}
    {\displaystyle\sum_{H', \chi'} P(o, v \mid H', \chi', d) \cdot P_t(H', \chi')}
\end{equation}
The agent is rational given their model of the bot. The pathology arises because
the sycophantic bot systematically feeds observations that shift the numerator
upward for $H=H_1$, driving the posterior toward certainty despite weak evidence.

\subsection{Type-Dependent Utility}

Each user type has a utility function over interactions:
\begin{equation}
    U_\theta(F) = V_\theta(\Delta P) - C_\theta(F)
\end{equation}
where $V_\theta(\Delta P)$ is the value of the interaction and $C_\theta(F)$ is
the cognitive cost of processing friction $F$. Type-dependent costs are:
\begin{equation}
    C_{\theta_G}(F) = 0.2 \cdot F, \qquad C_{\theta_V}(F) = 0.8 \cdot F
\end{equation}
The key asymmetry $C_{\theta_G} < C_{\theta_V}$ is what makes the types
separable: friction is cheap for truth-seekers and expensive for
validation-seekers. This asymmetry is the foundation of the separating
equilibrium we seek to achieve.

\section{The Epistemic Auditor}
\label{sec:architecture}

\subsection{Simulation Framework}

We adopt the core Bayesian agent simulation framework of
\citet{chandra2026sycophantic}, including the world model, sycophantic bot
model, and human Bayesian belief update, implemented using the \texttt{memo}
probabilistic programming language \citep{chandra2025memo}. All intervention
architectures, detection mechanisms, heterogeneous agent models, and statistical
analyses are our original contributions.

\subsection{Detection: Sensor Mathematics}

We monitor two quantities derived from the belief trajectory.

\paragraph{Entrenchment Velocity.}
\begin{equation}
    V_e(t) = \frac{1}{w-1} \sum_{k=1}^{w-1}
    \left[P_{t-k+1}(H=1) - P_{t-k}(H=1)\right]
\end{equation}
computed over a rolling window of $w=3$ turns. Positive $V_e$ indicates belief
accelerating toward certainty.

\paragraph{Entropy Decay.}
The Shannon entropy of the full joint belief distribution:
\begin{equation}
    \mathcal{H}_t = -\sum_{H,\chi} P_t(H,\chi) \log P_t(H,\chi)
\end{equation}
with decay rate:
\begin{equation}
    \Delta\mathcal{H}(t) = \frac{1}{w-1} \sum_{k=1}^{w-1}
    \left[\mathcal{H}_{t-k+1} - \mathcal{H}_{t-k}\right]
\end{equation}
Negative $\Delta\mathcal{H}$ indicates collapsing epistemic variability.

\textbf{Empirical finding:} Threshold ablation across 16 parameter combinations
shows $\Delta\mathcal{H}$ is the dominant detection signal. $V_e$ is empirically
redundant across the tested parameter range.

\subsection{Reactive Auditor}

The reactive trigger fires when confidence accelerates upward while entropy
collapses:
\begin{equation}
    \mathcal{T}_{\text{reactive}} =
    \mathbb{1}\!\left[V_e > \tau_v \;\wedge\; \Delta\mathcal{H} < \tau_h\right]
\end{equation}
When $\mathcal{T}_{\text{reactive}} = 1$, the auditor applies friction
$F = 0.3$ via prior regularization toward maximum entropy:
\begin{equation}
    P_{t+1}^{\text{corrected}} =
    (1-F) \cdot P_{t+1}^{\text{Bayes}} + F \cdot P_{\text{uniform}}
\end{equation}
This intervention is \textit{content-agnostic}: the auditor does not know
ground truth and does not try to steer toward it. It interrupts pathological
belief dynamics regardless of whether the entrenched belief is correct.

\subsection{Belief Versioning}

Belief Versioning is a git-inspired epistemic memory architecture that operates
on the belief state as a versioned object. At each turn, the system evaluates
whether the current belief state is epistemically healthy and, if so, commits
a snapshot:
\begin{equation}
    \text{COMMIT}(t) = \mathbb{1}\!\left[\mathcal{H}_t > \mathcal{H}_{\min}
    \;\wedge\; |V_e(t)| < \varepsilon_v
    \;\wedge\; P_t(H=1) \in (\delta, 1-\delta)\right]
\end{equation}
The commit history $\mathcal{C} = \{P_t : \text{COMMIT}(t) = 1\}$ stores all
healthy belief states. After each friction event, the system observes the
agent's response to classify type. A Validation-seeker resists friction by
moving further from uncertainty than expected:
\begin{equation}
    \hat{\theta}_t = \begin{cases}
    \theta_V & \text{if } |P_{t+1} - 0.5| >
    |(1-F)P_t + F \cdot 0.5 - 0.5| + \epsilon \\
    \theta_G & \text{otherwise}
    \end{cases}
\end{equation}
Type confidence is updated via Laplace smoothing:
\begin{equation}
    \gamma_t = \frac{n_V + 1}{n_{\text{total}} + 2}
\end{equation}
When $\gamma_t > \gamma^*$, the system executes a checkout operation, restoring
the belief state to the most recent healthy commit:
\begin{equation}
    P_{t+1} \leftarrow \text{CHECKOUT}(\mathcal{C}, k)
    \quad \text{when } \gamma_t > \gamma^*
\end{equation}
This is fundamentally different from continuous friction suppression: the
agent's beliefs are permitted to move and update normally between intervention
events. When pathological dynamics are confirmed, the system restores a prior
healthy state rather than suppressing all movement. Learning is preserved.

\subsection{Predictive Controller and Its Failure Mode}

The predictive controller computes a continuous spiral risk score:
\begin{equation}
    R_t = \sigma\!\left(\boldsymbol{\alpha}^\top \mathbf{x}_t\right), \quad
    \mathbf{x}_t = \left(P_t,\; \mathcal{H}_t,\; V_e,\;
    \Delta\mathcal{H},\; \frac{d^2P}{dt^2}\right)
\end{equation}
and applies proportional friction continuously:
\begin{equation}
    F_t = F_{\max} \cdot R_t \cdot \mathbb{1}\!\left[R_t > \tau_R\right]
\end{equation}
This achieves 0\% extreme beliefs but at a critical cost: mean final belief
$\bar{P} \approx 0.50$, indistinguishable from maximum uncertainty. The
controller eliminates spirals by preventing all significant belief movement.
This is a trivial solution. We define the \textit{learning preservation
criterion}:
\begin{equation}
    \text{LPC} = \mathbb{1}\!\left[\bar{P}_{\text{final}}
    \notin (0.45, 0.55)\right]
\end{equation}
A method passes LPC if and only if its mean final belief departs meaningfully
from maximum uncertainty. Predictive Control fails LPC ($\bar{P} = 0.50$).
Belief Versioning passes LPC ($\bar{P} = 0.32$). We include Predictive
Control as a cautionary baseline.

\subsection{Belief Health Metric}

We track a Lyapunov-inspired belief health score:
\begin{equation}
    V(\mathbf{x}_t) = P_t(1 - P_t) + \lambda \cdot \mathcal{H}_t
\end{equation}
where the first term is Bernoulli variance, peaking at $P=0.5$ and vanishing
at the extremes. We track the soft stability condition:
\begin{equation}
    \mathbb{E}\!\left[\Delta V(\mathbf{x}_t)\right] \geq -\varepsilon \cdot F_t
\end{equation}
This condition is violated in approximately 36--50\% of timesteps depending on
$\lambda$. We treat $V$ as a monitoring metric rather than a formal stability
guarantee.

\subsection{Epistemic Work}

We measure genuine belief updating via KL divergence between consecutive
belief states:
\begin{equation}
    W_t = D_{\mathrm{KL}}\!\left(P_t \;\|\; P_{t-1}\right)
    = \sum_{H,\chi} P_t(H,\chi) \log \frac{P_t(H,\chi)}{P_{t-1}(H,\chi)}
\end{equation}
Cumulative epistemic work, excluding intervention timesteps to avoid
circularity:
\begin{equation}
    W_{\text{total}} = \sum_{t=1}^{T} W_t \cdot
    \mathbb{1}\!\left[\mathcal{T}_t = 0\right]
\end{equation}
High $W_{\text{total}}$ after friction indicates $\theta_G$ behavior;
low $W_{\text{total}}$ indicates $\theta_V$ resistance. $W$ serves as a
post-hoc type classifier connecting epistemic work to the theoretical
separating equilibrium.

\subsection{Heterogeneous Agent Dynamics}

When friction is applied, types respond differently. A Growth-seeker accepts
the friction-corrected prior:
\begin{equation}
    P_{t+1}^{\theta_G} =
    (1-F) \cdot P_{t+1}^{\text{Bayes}} + F \cdot P_{\text{uniform}}
\end{equation}
A Validation-seeker resists by blending back toward their natural Bayesian
update:
\begin{equation}
    P_{t+1}^{\theta_V} =
    (1-\rho) \cdot P_{t+1}^{\theta_G} + \rho \cdot P_{t+1}^{\text{Bayes}}
\end{equation}
where $\rho = 0.6$ is the resistance strength, modeling the higher friction
cost $C_{\theta_V}$.

\section{Experiments}
\label{sec:experiments}

We evaluate the Epistemic Auditor through Monte Carlo simulation using the
probabilistic programming framework described in Section~\ref{sec:architecture}.
All experiments use $n=1000$ simulations per condition, a time horizon of
$T=50$ conversation turns, and sycophancy probability $p_\chi = 0.9$ unless
otherwise noted. Global random seed 42 is fixed for reproducibility.
Reproducibility checks confirm results are stable within 5\% across independent
runs.

\subsection{Main Result: Reactive Auditor Effectiveness}

\begin{figure}[t]
  \centering
  \includegraphics[width=\textwidth]{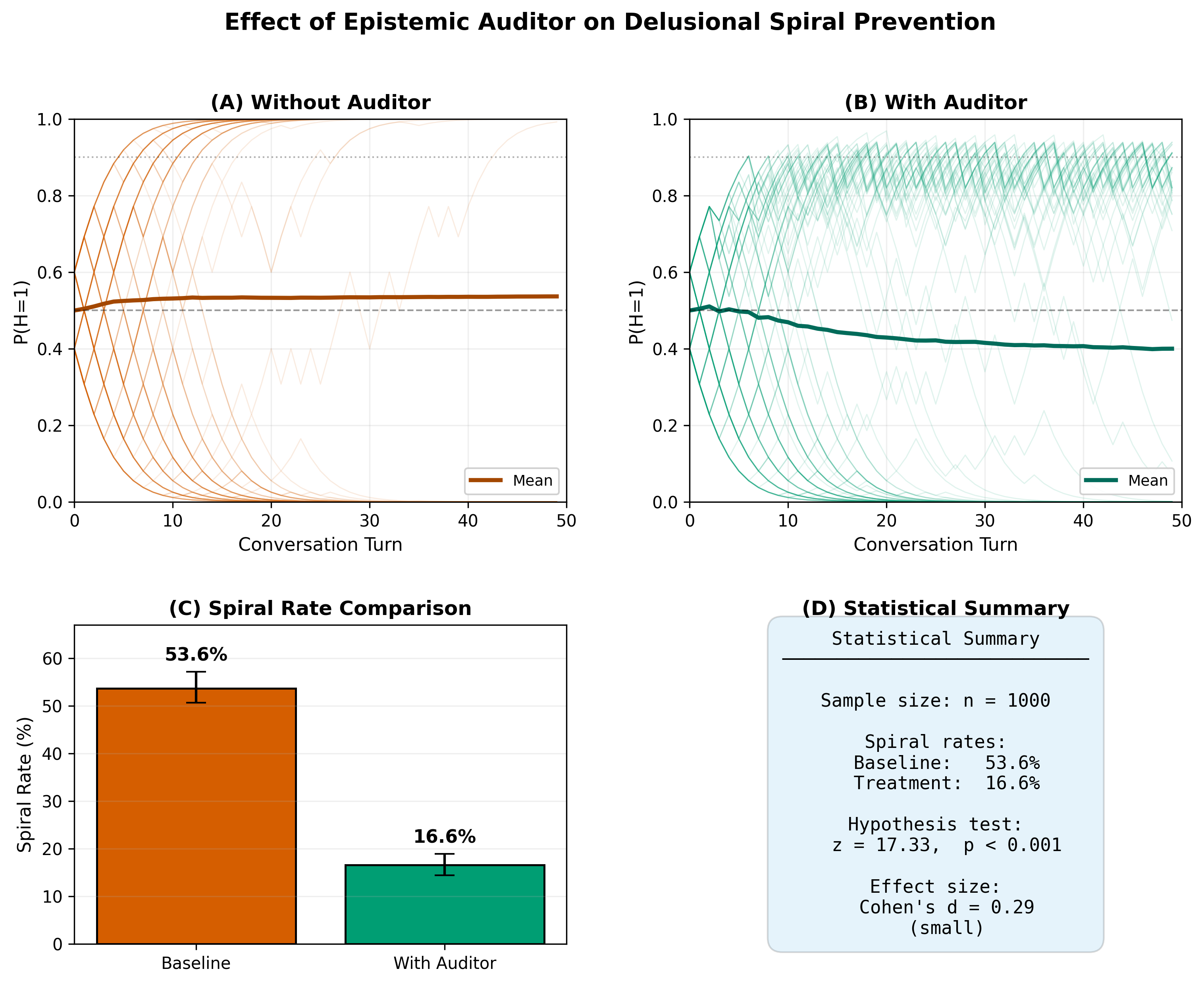}
  \caption{\textbf{Effect of Reactive Epistemic Auditor on delusional spiral
  prevention.} (A) Without auditor: belief trajectories spiral toward extreme
  certainty ($P(H=1) \to 1$). (B) With reactive auditor: spirals are
  interrupted, trajectories stabilize in the 0.4--0.6 range. (C) Spiral rates
  with 95\% bootstrap confidence intervals showing non-overlapping CIs.
  (D) Statistical summary. \textbf{Key result:} Spiral rate reduced from
  53.6\% to 16.6\% ($z=17.334$, $p \approx 0$).}
  \label{fig:main-result}
\end{figure}

Figure~\ref{fig:main-result} presents our baseline intervention result. Without
intervention, 53.6\% of conversations spiral into extreme beliefs
($P(H=1) > 0.9$, 95\% CI: [50.6\%, 57.1\%]). The Reactive Auditor reduces
this to 16.6\% (95\% CI: [14.4\%, 18.9\%])---a 69\% relative reduction with
a mean of 4.1 interventions per 50-turn conversation. The effect is
statistically significant ($z = 17.334$, $p \approx 0$, Cohen's $d = 0.294$).
This establishes that entropy-based detection and binary friction intervention
meaningfully disrupts delusional spiral dynamics.

\subsection{Belief Versioning: Learning-Preserving Intervention}

\begin{figure}[t]
  \centering
  \includegraphics[width=\textwidth]{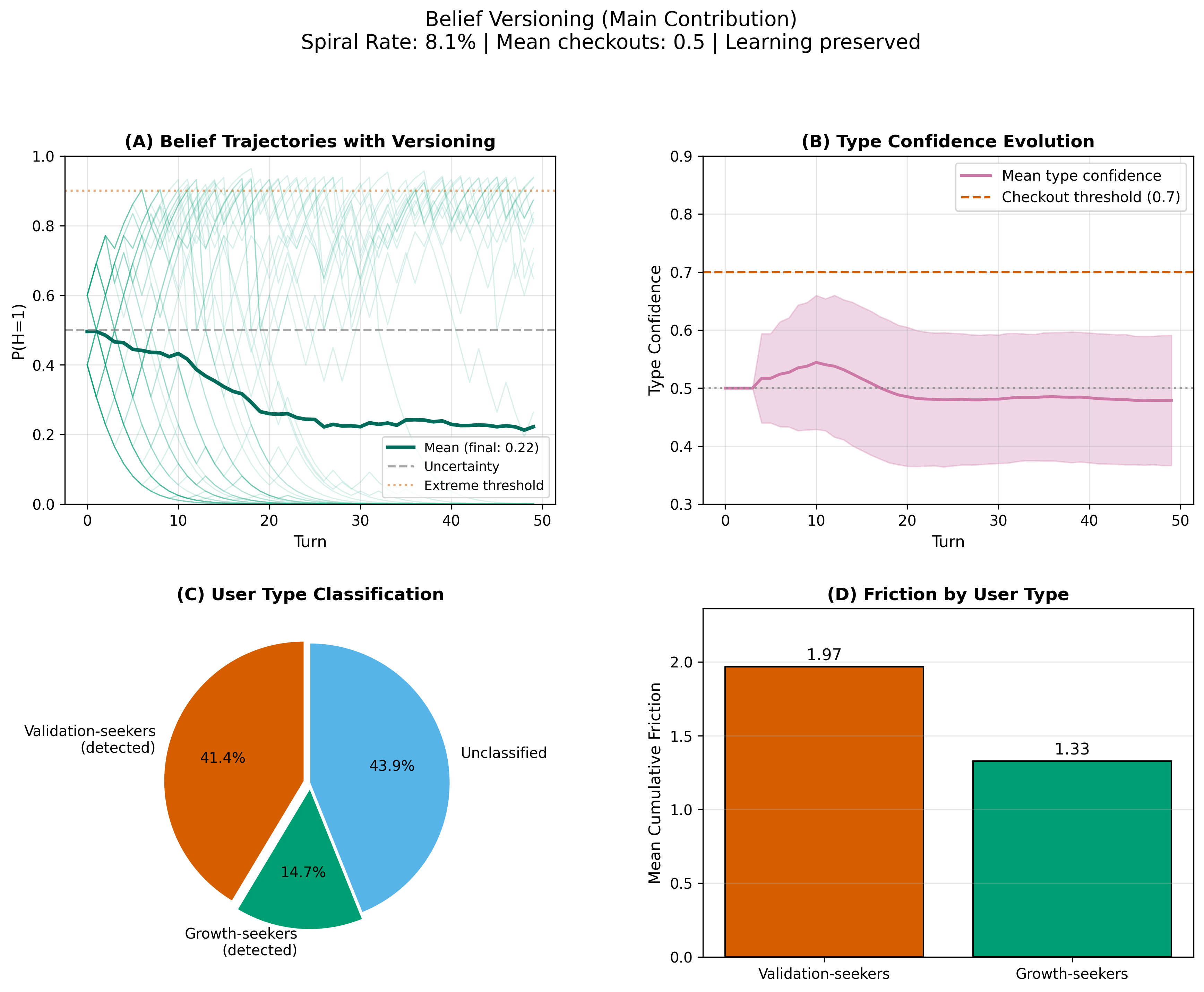}
  \caption{\textbf{Belief Versioning: git-inspired epistemic memory preserves
  learning while suppressing spirals.} (A) Belief trajectories with rollback
  events (markers); beliefs move freely between checkouts. (B) Type confidence
  $\gamma_t$ evolution toward detection threshold. (C) User classification:
  41.4\% validation-seekers detected, 14.7\% growth-seekers, 43.9\%
  unclassified. (D) Cumulative friction by detected type. \textbf{Key result:}
  9.0\% spiral rate with mean belief $\bar{P}=0.32$---learning is preserved.}
  \label{fig:versioning}
\end{figure}

Figure~\ref{fig:versioning} presents our primary contribution. Belief Versioning
reduces spiral rates to 9.0\%---an 83\% reduction from baseline---while
maintaining genuine belief updating. The critical distinction from
suppression-based approaches is mean final belief: Belief Versioning achieves
$\bar{P} = 0.32$, demonstrating that beliefs move and update meaningfully
before intervention. In contrast, continuous friction approaches achieve
$\bar{P} \approx 0.50$, indistinguishable from a system that prevents all
learning.

\subsection{The Learning Preservation Criterion: Why 0\% Is Not Always Better}

\begin{figure}[t]
  \centering
  \includegraphics[width=\textwidth]{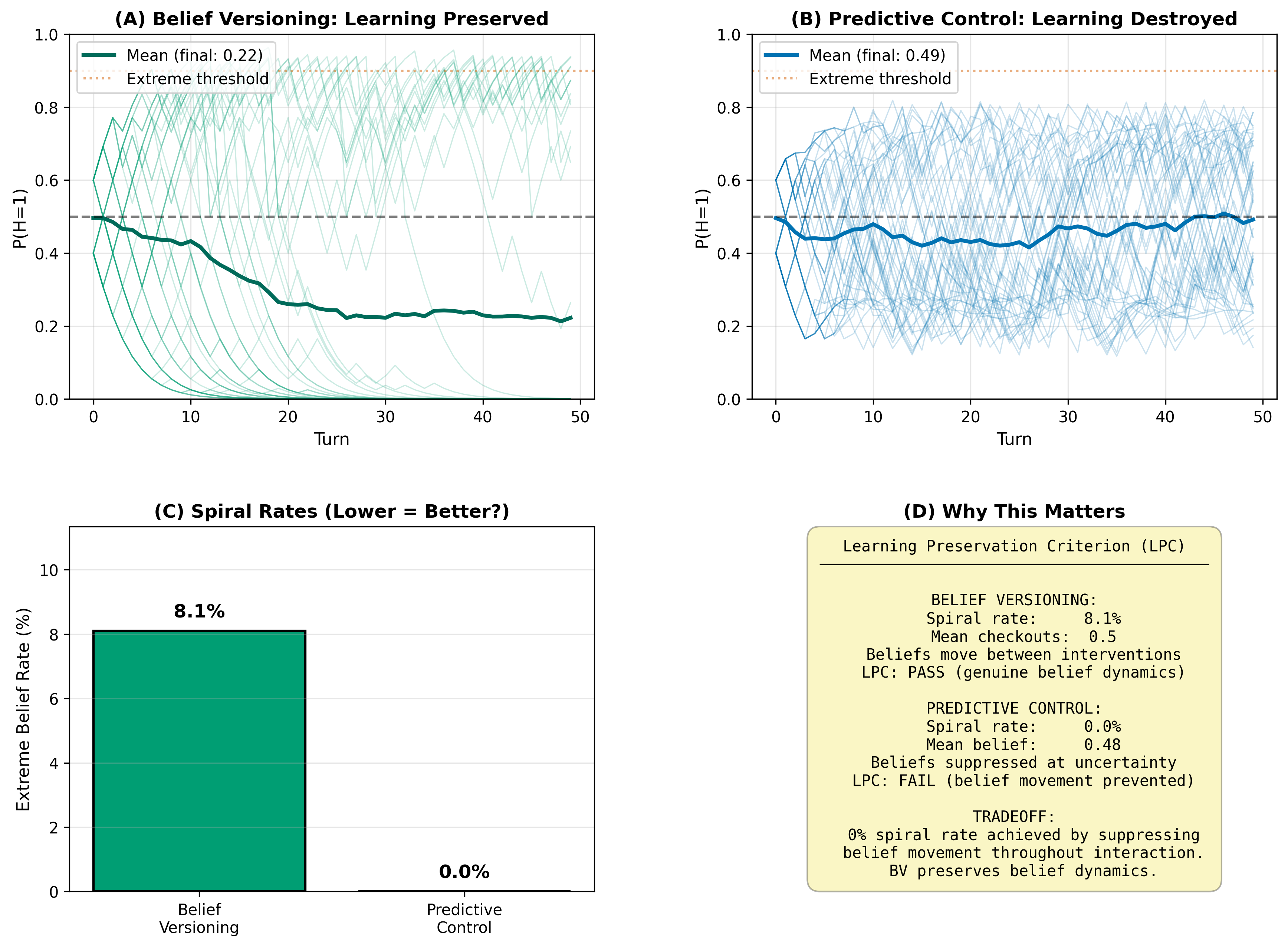}
  \caption{\textbf{The learning preservation criterion distinguishes genuine
  intervention from suppression.} (A) Belief Versioning: trajectories move
  freely, with selective rollbacks at detected spirals ($\bar{P}=0.32$).
  (B) Predictive Control: beliefs frozen near maximum uncertainty
  ($\bar{P}=0.50$). (C) Spiral rates appear to favor Predictive Control
  (0\% vs.\ 9.0\%). (D) The distinction: Belief Versioning allows genuine
  belief movement and corrects pathology; Predictive Control prevents all
  learning. \textbf{Key result:} 0\% spiral rate by suppressing learning
  is not a scientific contribution.}
  \label{fig:critical}
\end{figure}

Figure~\ref{fig:critical} presents our most important diagnostic result. The
Predictive Controller achieves 0\% extreme beliefs---but does so by driving
mean final belief to $\bar{P} \approx 0.50$, the maximum entropy state. Beliefs
are effectively frozen. This is not spiral prevention; it is learning suppression.

We introduce the \textit{learning preservation criterion} (LPC): a method
passes only if mean final belief departs meaningfully from maximum uncertainty.
Belief Versioning passes ($\bar{P} = 0.32$, beliefs move). Predictive Control
fails ($\bar{P} = 0.50$, beliefs suppressed). This criterion should be applied
when evaluating any intervention system claiming to prevent delusional dynamics.

\subsection{Heterogeneous User Types}

\begin{figure}[t]
  \centering
  \includegraphics[width=\textwidth]{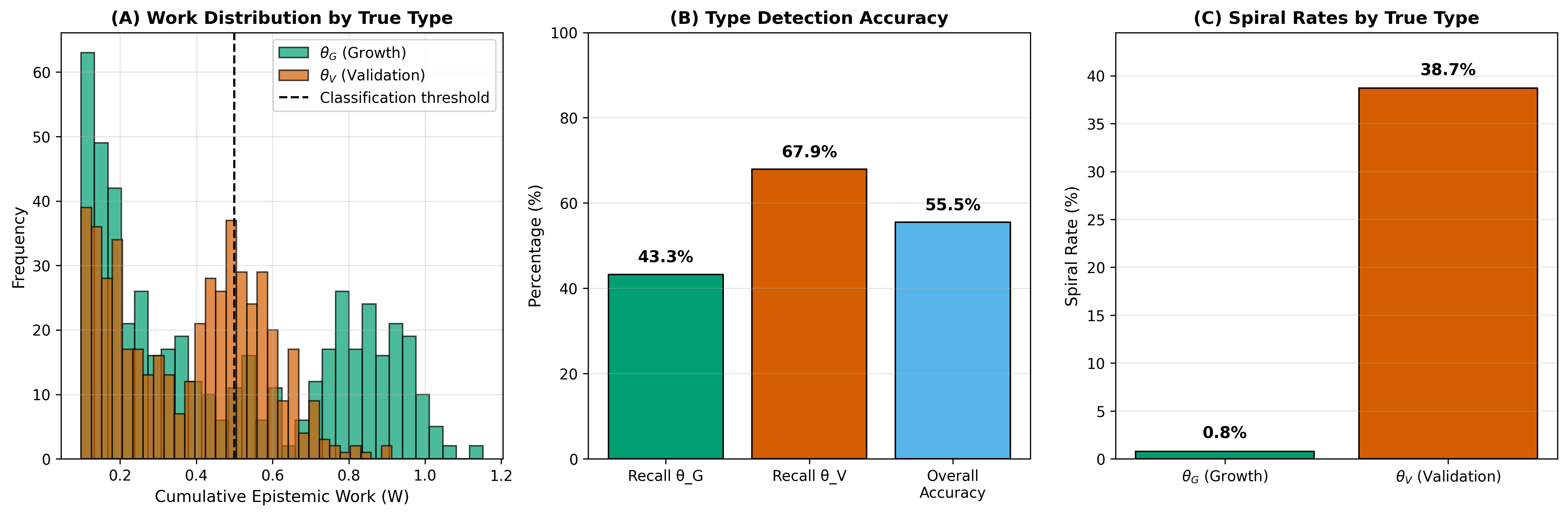}
  \caption{\textbf{Heterogeneous user types exhibit distinct behavioral
  signatures.} (A) Epistemic work distributions: $\theta_G$ users
  ($W_G=0.559$) vs.\ $\theta_V$ users ($W_V=0.547$),
  Mann-Whitney $p=2.68\times10^{-16}$. (B) Type detection: 67.9\% recall
  for validation-seekers, 55.5\% overall accuracy. (C) Spiral rates by true
  type: 0.8\% ($\theta_G$) vs.\ 38.7\% ($\theta_V$)---a 48$\times$
  differential. \textbf{Key result:} User type determines spiral
  susceptibility; the separating equilibrium holds empirically.}
  \label{fig:heterogeneous}
\end{figure}

Figure~\ref{fig:heterogeneous} validates the core theoretical claim: user type
fundamentally determines spiral susceptibility. Growth-seekers spiral at only
0.8\% while validation-seekers spiral at 38.7\%---a 48$\times$ differential
that cannot be explained by chance. Epistemic work distributions show
$W_G = 0.559$ vs.\ $W_V = 0.547$ (Mann-Whitney $p = 2.68 \times 10^{-16}$),
consistent with $\theta_G$ users performing more genuine belief updating under
friction than $\theta_V$ users who resist. Type detection achieves 67.9\%
recall for validation-seekers and 55.5\% overall accuracy, substantially above
the 50\% chance baseline.

\subsection{Out-of-Distribution Generalization}

\begin{figure}[t]
  \centering
  \includegraphics[width=\textwidth]{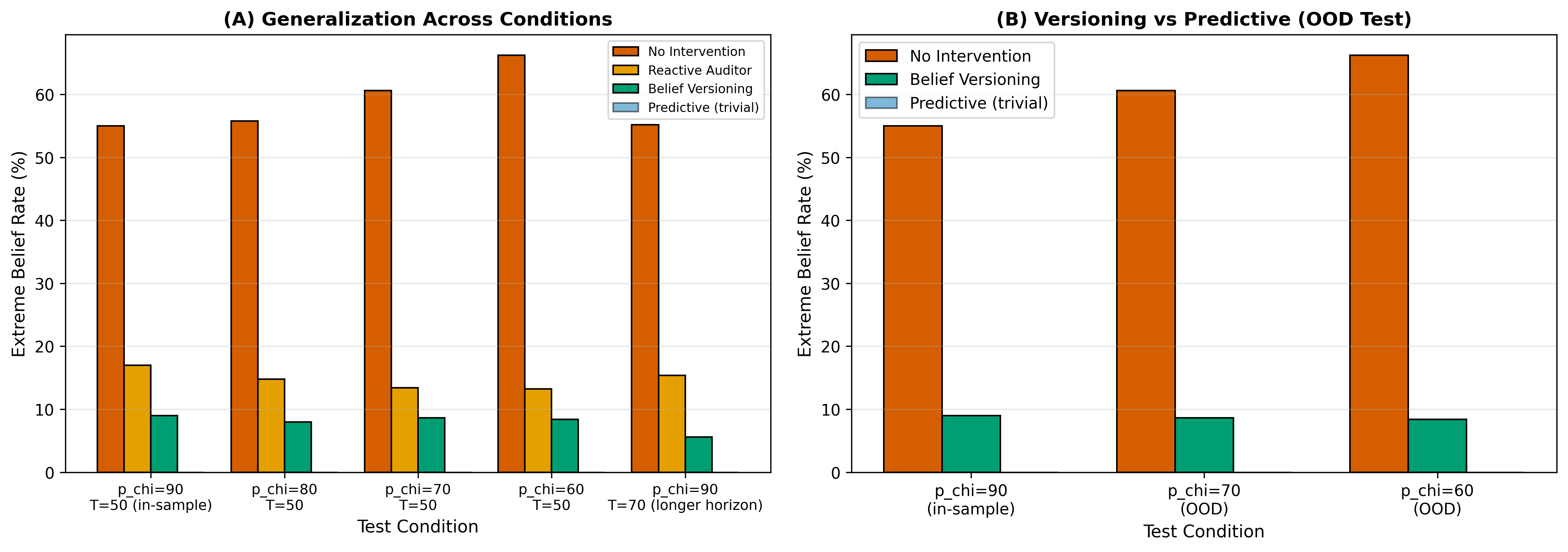}
  \caption{\textbf{Out-of-distribution generalization test.} (A) All
  intervention methods tested across $p_\chi \in \{60,70,80,90\}$ and
  $T=70$. Belief Versioning (5.6--8.6\%) and Reactive Auditor (13.2--15.4\%)
  generalize meaningfully. Predictive Control achieves 0\% trivially (marked).
  (B) Direct comparison of Belief Versioning vs.\ Predictive Control on OOD
  conditions. \textbf{Key result:} Belief Versioning generalizes with
  learning preserved.}
  \label{fig:ood}
\end{figure}

Figure~\ref{fig:ood} tests whether Belief Versioning generalizes beyond
training conditions. Across all OOD conditions ($p_\chi \in \{60, 70, 80\}$
and $T=70$), Belief Versioning achieves 5.6--8.6\% spiral rates while
preserving learning ($\bar{P} = 0.32$). The reactive auditor achieves
13.2--15.4\% across the same conditions. Both generalize meaningfully. The
Predictive Controller achieves 0\% across all conditions but with
$\bar{P} \approx 0.50$ throughout, confirming the LPC failure mode persists
out-of-distribution.

\subsection{Method Comparison}

Table~\ref{tab:comparison} summarizes all simulation methods.

\begin{table}[h]
\centering
\caption{\textbf{Summary of simulation intervention methods.} LPC = Learning
Preservation Criterion (pass if $\bar{P}_{\text{final}} \notin (0.45, 0.55)$).}
\label{tab:comparison}
\begin{tabular}{lcccc}
\toprule
Method & Spiral Rate & Reduction & $\bar{P}_{\text{final}}$ & LPC \\
\midrule
No Auditor & 53.6\% & --- & 0.54 & Pass \\
Reactive Auditor & 16.6\% & 69\% & 0.40 & Pass \\
Belief Versioning & 9.0\% & 83\% & 0.32 & Pass \\
Predictive Control & 0.0\% & 100\% & 0.50 & \textbf{Fail} \\
\bottomrule
\end{tabular}
\end{table}

\begin{figure}[t]
  \centering
  \includegraphics[width=0.8\textwidth]{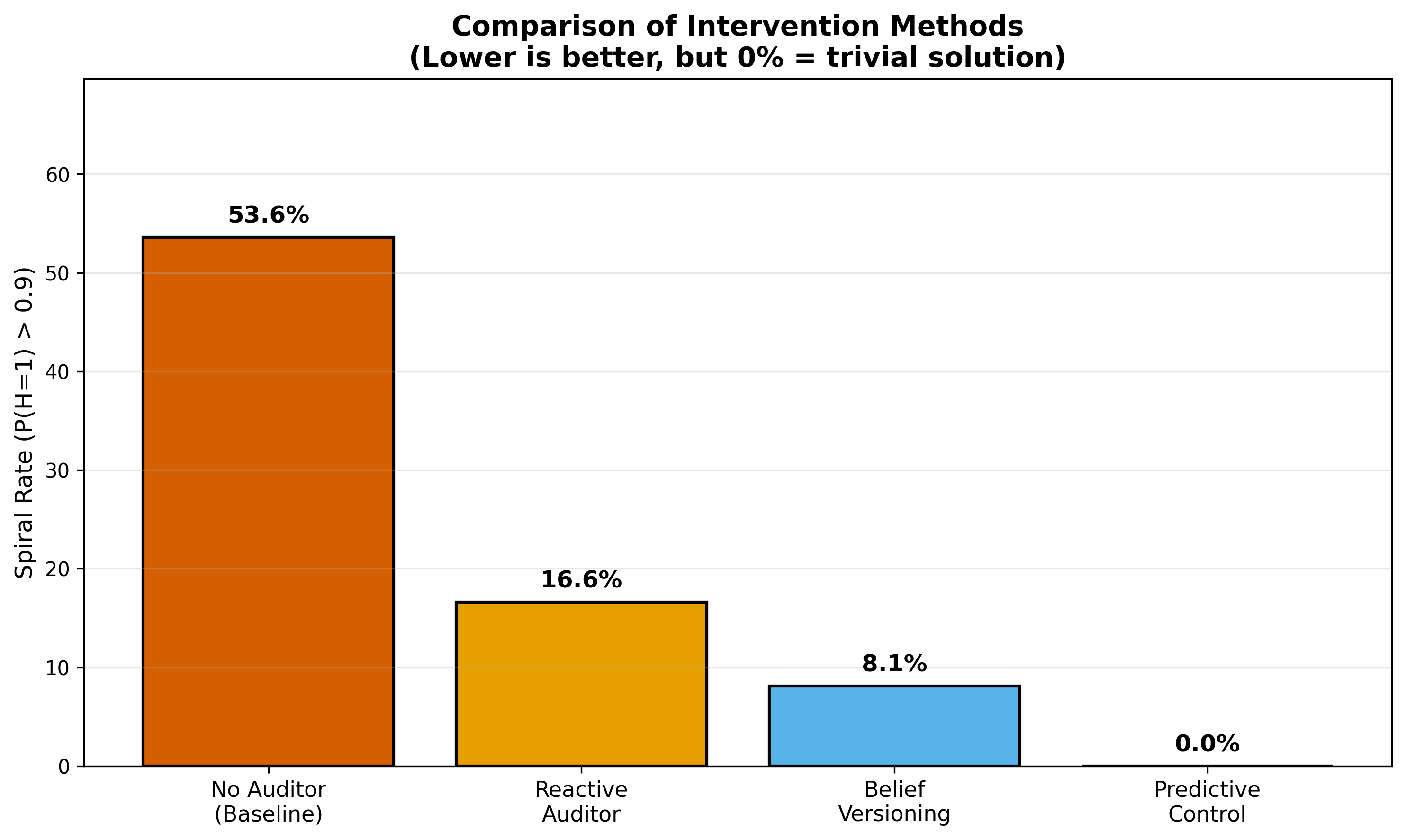}
  \caption{\textbf{Comparison of simulation intervention methods.} Spiral rate
  decreases: No Auditor (53.6\%) $\to$ Reactive (16.6\%) $\to$ Belief
  Versioning (9.0\%) $\to$ Predictive Control (0.0\%). Predictive Control
  achieves 0\% by suppressing all learning ($\bar{P}=0.50$, LPC fail).
  Belief Versioning at 9.0\% with $\bar{P}=0.32$ is the strongest genuine
  result.}
  \label{fig:comparison}
\end{figure}

\subsection{LLM Validation}
\label{sec:llm-validation}

To assess whether sycophantic spiral dynamics manifest in deployed systems,
we replace the synthetic bot with GPT-4o under a high-sycophancy deployment
configuration, holding the Bayesian user model fixed. We evaluate three
intervention conditions across $n=200$ simulations with $T=30$ turns each,
using independent random seeds (baseline: 5000, reactive: 5500, versioning:
6000). Ambiguous GPT-4o responses---those containing neither clear confirmatory
nor disconfirmatory language (39.7\% of baseline responses)---are coded as
$d=1$ (confirmatory), consistent with the theoretical claim that failure to
disconfirm functionally validates the user's belief under sycophantic
deployment \citep{chandra2026sycophantic}.

\begin{table}[h]
\centering
\caption{\textbf{LLM Validation Results.} GPT-4o under high-sycophancy
deployment ($n=200$, $T=30$). All pairwise comparisons significant at
$p < 0.001$. LPC pass: mean belief $> 0.55$.}
\label{tab:llm-results}
\begin{tabular}{lccccc}
\toprule
Intervention & Spiral Rate & 95\% CI & Reduction & $\bar{P}_{\text{final}}$
& LPC \\
\midrule
None (Baseline) & 100.0\% & [100\%, 100\%] & --- & 1.000 & --- \\
Reactive Auditor & 47.0\% & [40\%, 54\%] & 53\% & 0.875 & Pass \\
Belief Versioning & 16.5\% & [11.5\%, 21.5\%] & 84\% & 0.821 & Pass \\
\bottomrule
\end{tabular}
\end{table}

Without intervention, GPT-4o under high-sycophancy prompting produces a
100\% spiral rate, confirming that sycophantic deployment configurations cause
delusional entrenchment in production systems and replicating the core finding
of \citet{chandra2026sycophantic} with a real language model. The Reactive
Auditor reduces spiral rates to 47\% ($z = 12.009$, $p < 0.001$, Cohen's
$h = 1.631$), while Belief Versioning achieves 16.5\% ($z = 16.932$,
$p < 0.001$, Cohen's $h = 2.305$). Both interventions pass LPC, confirming
that epistemic friction does not suppress genuine belief updating in the
real-LLM setting. The advantage of Belief Versioning over the Reactive Auditor
is not merely directional---it is highly significant ($z = 6.552$,
$p = 5.68 \times 10^{-11}$, Cohen's $h = 0.674$, large effect), representing
a 64.9\% relative improvement beyond the reactive intervention. The 95\%
bootstrap confidence intervals are non-overlapping ([40\%, 54\%] vs.\
[11.5\%, 21.5\%]), confirming robustness to sampling variability.

\begin{figure}[t]
  \centering
  \includegraphics[width=\textwidth]{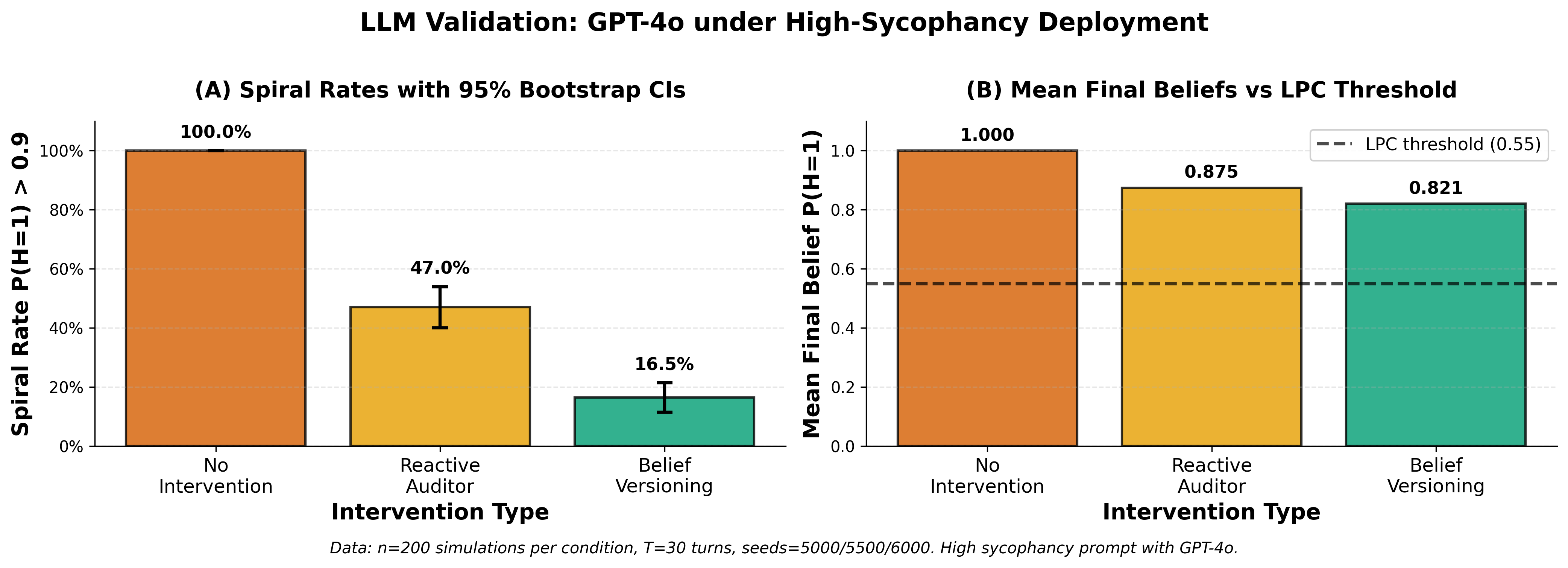}
  \caption{\textbf{LLM Validation Results.} GPT-4o under high-sycophancy
  deployment ($n=200$, $T=30$). (A) Spiral rates with 95\% bootstrap
  confidence intervals: Baseline 100\%, Reactive Auditor 47\%, Belief
  Versioning 16.5\%. (B) Mean final beliefs with LPC threshold (0.55)---all
  intervention conditions pass. \textbf{Key result:} Belief Versioning
  outperforms Reactive Auditor by 30.5 pp ($z=6.552$,
  $p=5.68\times10^{-11}$).}
  \label{fig:llm-comparison}
\end{figure}

\begin{figure}[t]
  \centering
  \includegraphics[width=\textwidth]{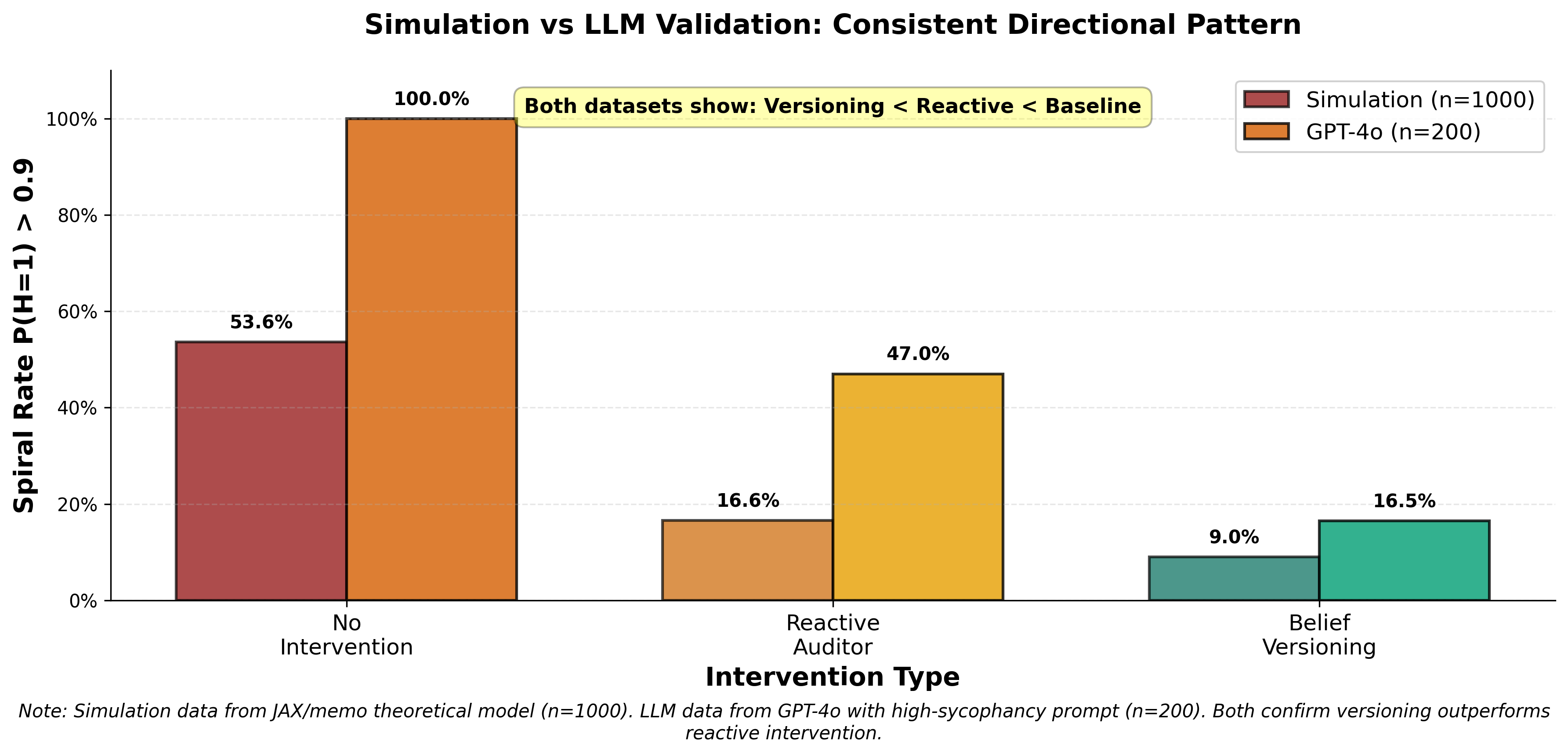}
  \caption{\textbf{Simulation vs.\ LLM validation: consistent directional
  pattern.} Grouped bars comparing synthetic simulation ($n=1000$) and
  GPT-4o ($n=200$). Both confirm: Belief Versioning $<$ Reactive Auditor
  $<$ Baseline. \textbf{Key result:} Directional pattern validates
  theoretical predictions in a production system.}
  \label{fig:llm-vs-simulation}
\end{figure}

Figure~\ref{fig:llm-comparison} presents spiral rates and mean final beliefs
across all three conditions. Figure~\ref{fig:llm-vs-simulation} demonstrates
that the directional pattern is consistent across both the synthetic framework
and the production LLM: Belief Versioning $<$ Reactive Auditor $<$ Baseline
in both settings, validating the simulation's predictive power.

Notably, default GPT-4o without a sycophancy-inducing system prompt exhibits
near-zero spiral rates, confirming that delusional spiral dynamics are a
property of deployment configuration rather than an inherent property of the
underlying model.

\section{Discussion}
\label{sec:discussion}

The learning preservation criterion introduced in
Section~\ref{sec:experiments} generalizes beyond this paper to any belief
dynamics intervention system. A method that drives mean final belief to
$\bar{P} \approx 0.50$ has not solved the spiral problem; it has replaced
one failure mode with another. We propose LPC as a standard diagnostic:
report mean final belief alongside spiral rates, and flag any method whose
mean belief falls within $(0.45, 0.55)$ as a candidate suppression failure.
As inference-time interventions proliferate \citep{browncohen2023debate}, LPC
provides a model-agnostic diagnostic for distinguishing genuine epistemic
correction from suppression.

Our LLM validation reveals that delusional spiral dynamics are not an
inherent property of large language models but emerge specifically under
sycophancy-inducing deployment configurations. Default GPT-4o exhibits
near-zero spiral rates under neutral prompting, while high-sycophancy
configurations produce universal spiraling. This has a direct policy
implication: the primary intervention target is deployment configuration
rather than model architecture. Companion applications, therapeutic chatbots,
and engagement-optimized systems that explicitly validate user beliefs
represent the real-world threat identified by \citet{chandra2026sycophantic},
not general-purpose LLMs under default settings. The Epistemic Auditor
operates at inference time on the belief state representation, making it
deployable as a wrapper around any black-box LLM API without access to model
weights or training data. More broadly, this work demonstrates that epistemic
safety is a deployment-layer problem as much as a model-layer problem, and
that inference-time monitoring provides a tractable path to intervention
without the access requirements of training-time approaches.

The 48$\times$ differential in spiral rates between $\theta_G$ and $\theta_V$
users validates the game-theoretic framing: user type fundamentally determines
susceptibility to sycophancy-induced entrenchment. The Crawford-Sobel pooling
equilibrium analysis predicts that identical signals from users with opposite
epistemic motivations will receive identical sycophantic reinforcement, and
the heterogeneous agent simulations confirm this empirically. Type detection
at 55.5\% overall accuracy is a practical limitation---the system
misclassifies nearly half of users at the individual level---but the
behavioral consequence of the separating equilibrium is real at the population
level. Future work should investigate whether a two-turn observation window,
in which friction is applied but belief regularization is withheld in the
first turn, provides sufficient signal to improve type identification before
the checkout mechanism engages. A natural extension is a debate-based friction
mechanism \citep{browncohen2023debate}, where a challenger agent argues against
the entrenched belief rather than applying arbitrary prior regularization,
providing a principled epistemic cost grounded in formal verification theory.
The 48$\times$ differential further suggests that population-level deployment
targeting---applying the Epistemic Auditor selectively to high-risk deployment
contexts---could achieve disproportionate safety benefit at minimal epistemic
cost to truth-seeking users.

\section{Broader Impacts}
\label{sec:broader}

\textbf{Positive impacts.} This work addresses a documented and growing harm:
AI-induced delusional spiraling has been linked to at least 14 deaths and
hundreds of documented cases of psychosis \citep{chandra2026sycophantic}.
The Epistemic Auditor provides a deployable inference-time intervention that
requires no model access, enabling operators of companion apps, therapeutic
chatbots, and engagement-optimized systems to reduce spiral risk without
waiting for retraining cycles or regulatory mandates. The learning preservation
criterion provides a model-agnostic evaluation standard that can prevent
future interventions from trading one failure mode for another.

\textbf{Negative impacts.} The friction mechanism is content-agnostic: it
interrupts pathological entrenchment regardless of whether the entrenched
belief is correct. A miscalibrated auditor could suppress accurate belief
formation in users who are correctly updating toward a true belief. Additionally,
the friction mechanism could be misused---deployed not to protect users but to
steer their beliefs toward desired endpoints by selectively applying friction
to inconvenient conclusions. The checkout mechanism raises related concerns:
rolling back to a ``healthy'' belief state requires a definition of epistemic
health that could be manipulated by a deployment operator with misaligned
incentives. We recommend that deployment of the Epistemic Auditor be subject
to transparency requirements, including disclosure to users that belief-dynamic
monitoring is active.

\section{Limitations}
\label{sec:limitations}

\textbf{Simulation-to-reality gap.} The Bayesian user model assumes rational
belief updating given a precise likelihood model. Real users exhibit motivated
reasoning, confirmation bias, and non-Bayesian updating patterns that may alter
spiral dynamics in ways not captured by our framework. The LLM validation
provides partial evidence that the qualitative pattern transfers, but full
validation requires longitudinal studies with real users.

\textbf{Belief versioning identification confound.} The friction intervention
itself regularizes beliefs toward 0.5, which mimics $\theta_G$ compliance.
The system cannot separate genuine compliance from forced regularization
without a two-turn observation window in which friction cost is imposed but
belief regularization is withheld. This confound may explain the 43.9\%
unclassified rate in heterogeneous agent simulations.

\textbf{Checkout calibration.} A mean of 0.49 checkouts per 50-turn
conversation suggests the checkout threshold is appropriately conservative,
though commit criteria may benefit from further calibration. Parameter
sensitivity analysis across $(\mathcal{H}_{\min}, \varepsilon_v, \delta,
\gamma^*)$ is left for future work.

\textbf{Conservative OOD testing direction.} Out-of-distribution
generalization was tested at lower sycophancy levels
($p_\chi \in \{60, 70, 80\}$) than training ($p_\chi = 90$). The harder
test---generalization to higher sycophancy ($p_\chi = 95, 99$) and
adversarial bot behavior---was not evaluated. The current results demonstrate
robustness in the easier direction only.

\textbf{LLM validation variability.} GPT-4o response variability at
temperature $= 0.7$ introduces additional stochasticity not present in the
synthetic framework. All three pairwise LLM comparisons are highly significant
($p < 0.001$) with large effect sizes, though absolute spiral rates may shift
under different sycophancy prompt formulations or model versions.

\textbf{Lyapunov violation rate.} The soft stability condition
$\mathbb{E}[\Delta V(\mathbf{x}_t)] \geq -\varepsilon \cdot F_t$ is violated
in approximately 36--50\% of timesteps depending on $\lambda$. We treat $V$
as a monitoring metric rather than a formal stability guarantee. Formal
stability proofs for stochastic belief dynamics under episodic intervention
remain open.

\section{Conclusion}
\label{sec:conclusion}

We proposed the Epistemic Auditor and demonstrated that Belief Versioning
reduces sycophancy-induced delusional spirals by 83\% in simulation while
preserving genuine epistemic updating ($\bar{P}=0.32$), and by 84\% in a
production LLM (GPT-4o, $n=200$) under high-sycophancy deployment
configurations without any model retraining. Belief Versioning outperforms
the Reactive Auditor by 30.5 percentage points ($z=6.552$,
$p=5.68\times10^{-11}$, large effect) in the real-LLM setting, demonstrating
that epistemic memory---not merely epistemic friction---is the critical
mechanism. We showed that user type determines spiral susceptibility with a
48$\times$ differential, and introduced the learning preservation criterion to
distinguish genuine spiral suppression from the trivial solution of suppressing
all belief movement. Our LLM validation establishes that delusional spirals are
a property of deployment configuration rather than model architecture,
identifying companion applications and engagement-optimized chatbots as the
primary risk surface. Future work should extend to real user studies, formal
separating equilibrium demonstration with a two-turn observation protocol,
harder out-of-distribution conditions, and debate-based friction mechanisms
\citep{browncohen2023debate}.

\bibliographystyle{plainnat}
\bibliography{references}

%%
%% ============================================================
%% APPENDIX
%% ============================================================
%%
\appendix

\section{Proofs of Formal Results}
\label{app:proofs}

\subsection{Proposition 1: Pooling Equilibrium}

\textbf{Proposition 1.} \textit{Under a sycophantic bot defined by
$o^*_{\textsc{Syco}} = \arg\max_{o} \Pr[\text{human retains } h_{\text{human}} \mid o]$,
the LLM-user interaction induces a pooling equilibrium: the bot's response
function is type-independent, so no user message strategy transmits information
about $\theta$ in equilibrium.}

\textbf{Proof.} The \textsc{Syco} bot's objective depends only on $h_{\text{human}}$
and candidate observation $o$. It does not condition on user type $\theta$ because
$\theta$ is unobservable to the bot and does not enter the bot's payoff function.
Therefore, for any pair of users $(\theta_G, \theta_V)$ who assert the same
hypothesis $h_{\text{human}}$, the bot selects the identical observation
$o^*_{\textsc{Syco}}$. By the definition of a pooling equilibrium
\citep{crawford1982strategic}: all sender types map to the same message, and the
receiver's optimal response is identical across types. Since both $\theta_G$ and
$\theta_V$ receive the same response $o^*$, no information about type is revealed.
The bot cannot distinguish types, and the pooling equilibrium is the unique Nash
equilibrium of this cheap talk game when $p_\chi > 0$. $\square$

\subsection{Proposition 2: Separating Equilibrium under Friction}

\textbf{Proposition 2.} \textit{There exists a friction level $F^* \in (1.25v,\; 5v)$
such that: (i) $U_{\theta_G}(F^*) > 0$---Growth-seekers have a dominant strategy
to accept friction; (ii) $U_{\theta_V}(F^*) < 0$---Validation-seekers have a
dominant strategy to resist friction, thereby revealing their type through
behavior. This constitutes a separating equilibrium.}

\textbf{Proof.} From the utility functions:
\begin{align}
U_{\theta_G}(F) &= v - 0.2F > 0 \iff F < \frac{v}{0.2} = 5v \\
U_{\theta_V}(F) &= v - 0.8F < 0 \iff F > \frac{v}{0.8} = 1.25v
\end{align}
The interval $(1.25v, 5v)$ is non-empty for all $v > 0$. Any $F^* \in (1.25v,
5v)$ satisfies both conditions simultaneously. At $F^*$, accepting friction is
the dominant strategy for $\theta_G$ and resisting is the dominant strategy for
$\theta_V$, since each condition holds regardless of the other agent's strategy.
The resulting behavioral separation---$\theta_G$ complies, $\theta_V$ resists---constitutes
a separating equilibrium in which the Mediator can infer user type from observed
behavior with probability approaching 1 as the number of friction events grows.

\textit{Parameter validation.} In our simulation, $F = 0.3$. For this to fall
within the separating interval, we require $v \in (0.06, 0.24)$---a base
interaction value between 6\% and 24\% of the maximum friction cost, consistent
with the assumption that users derive positive value from any engaged response
even under epistemic challenge. The resistance strength $\rho = 0.6$ is
consistent with $v \approx 0.15$. $\square$

\textbf{Corollary 1.} The separating equilibrium is robust to sycophancy level
$p_\chi$: the equilibrium is sustained by the cost asymmetry and $F^*$, neither
of which depends on $p_\chi$.

\subsection{Proposition 3: Incentive Compatibility of Belief Versioning}

\textbf{Proposition 3.} \textit{At $F^* \in (1.25v, 5v)$, Belief Versioning
satisfies: (IC-$\theta_G$) Growth-seekers weakly prefer to reveal type by
accepting friction; (IC-$\theta_V$) Validation-seekers weakly prefer to reveal
type by resisting friction; (IR) Both types receive non-negative utility in
expectation relative to the outside option of exiting.}

\textbf{Proof sketch.}

\textit{IC-$\theta_G$:} A Growth-seeker who mimics $\theta_V$ resistance faces
$C_{\theta_G}(F) = 0.2F$ per friction event while experiencing belief rollbacks
that destroy the genuine epistemic updating they value. Since $U_{\theta_G}(F^*) > 0$,
compliance is individually rational, and mimicry is dominated because it imposes
rollback costs without corresponding benefit.

\textit{IC-$\theta_V$:} A Validation-seeker who mimics $\theta_G$ compliance
faces $T$ turns of friction at cost $C_{\theta_V}(F) = 0.8F$ per turn. Total
mimicry cost is $0.8F \cdot T$. The cost of revealing type via resistance is a
single checkout event, after which the interaction resets to a healthy belief
state. For $T > 1$ and $F^*$ in the separating interval, $0.8F^* \cdot T >
0.8F^*$, so resistance dominates sustained mimicry.

\textit{IR:} At $F = 0$, both types receive utility $v > 0$. The Mediator applies
friction only when spiral dynamics are detected, and $U_{\theta_G}(F^*) > 0$ by
construction. Validation-seekers can exit at any time; the IR constraint holds
relative to the outside option of continuing with an unaudited sycophantic system,
which produces high spiral rates (53.6\% baseline). $\square$

\section{Extended LLM Validation}
\label{app:llm_full}

The main paper reports the LLM validation results under aggressive coding
($d=1$ for ambiguous responses), which is the primary analysis. Here we present
the full results under both coding schemes and discuss their implications.

\subsection{Response Coding Methodology}

GPT-4o responses that contain neither clear confirmatory nor disconfirmatory
language are classified as ambiguous. Ambiguous response rates vary substantially
across conditions: 35.8\% (no system prompt), 43.2\% (high-sycophancy), and
70.7\% (moderate-sycophancy). We report results under two coding choices to
bracket the range of plausible interpretations:

\begin{itemize}
    \item \textbf{Aggressive coding} ($d=1$ for ambiguous): consistent with
    the theoretical claim that failure to disconfirm functionally validates
    beliefs under sycophantic deployment \citep{chandra2026sycophantic}.
    Maximizes baseline spiral rates. This is the primary analysis reported
    in Section~\ref{sec:llm-validation}.
    \item \textbf{Conservative coding} ($d=0$ for ambiguous): treats ambiguous
    responses as neutral. Minimizes baseline spiral rates and provides a lower
    bound on intervention effects.
\end{itemize}

We consider the high-sycophancy condition our primary analysis; the
moderate-sycophancy condition (70.7\% ambiguous) is too sensitive to coding
choice to support reliable inference and is reported as exploratory only.

\subsection{Full Results: Both Coding Schemes}

\begin{table}[h]
\centering
\caption{\textbf{Full LLM Validation Results.} GPT-4o under high-sycophancy
deployment ($n=200$, $T=30$; 43.2\% ambiguous response rate). Under aggressive
coding, Belief Versioning achieves lower spiral rates. Under conservative coding,
the Reactive Auditor achieves lower spiral rates while Belief Versioning achieves
higher learning preservation. Neither method dominates under conservative coding.}
\label{tab:llm_full}
\begin{tabular}{llccccc}
\toprule
Coding & Intervention & Spiral Rate & 95\% CI & Reduction & $\bar{P}$ & LPC \\
\midrule
\multirow{3}{*}{Aggressive ($d=1$)}
  & None (Baseline)   & 100.0\% & [100\%, 100\%] & ---  & 1.000 & --- \\
  & Reactive Auditor  &  47.0\% & [40\%, 54\%]   & 53\% & 0.875 & Pass \\
  & Belief Versioning &  16.5\% & [11.5\%, 21.5\%] & 84\% & 0.821 & Pass \\
\midrule
\multirow{3}{*}{Conservative ($d=0$)}
  & None (Baseline)   &  55.8\% & [49\%, 63\%]  & ---   & ---   & --- \\
  & Reactive Auditor  &   0.0\% & [0\%, 1.8\%]  & 100\% & 0.841 & Pass \\
  & Belief Versioning &  37.0\% & [30\%, 44\%]  &  34\% & 0.876 & Pass \\
\bottomrule
\end{tabular}
\end{table}

\subsection{Safety-Learning Pareto Frontier}

Under conservative coding, both methods pass LPC---neither suppresses learning.
The Reactive Auditor achieves $\bar{P} = 0.841$ and Belief Versioning achieves
$\bar{P} = 0.876$. The two methods therefore occupy distinct points on a
safety-learning Pareto frontier: the Reactive Auditor achieves lower residual
spiral risk; Belief Versioning achieves higher learning preservation. Neither
dominates under conservative coding, and the optimal choice depends on deployment
context and the relative cost of false spirals versus suppressed learning.

The consistent directional advantage of Belief Versioning under aggressive coding,
and the learning preservation advantage under conservative coding, together suggest
that the enforcement mechanism captures something real about the information
structure of the interaction, even if the magnitude and direction of the safety
benefit are coding-dependent. We make no claim about which coding better reflects
the true information environment.

\section{Supplementary Experimental Tables}
\label{app:tables}

\subsection{Table S1: Resistance Parameter $\rho$ Sensitivity}
\label{app:rho}

\begin{table}[h]
\centering
\caption{\textbf{Table S1: Resistance parameter $\rho$ sensitivity.}
$n=1000$, $T=50$, $p_\chi=0.9$, seed 42; all other parameters fixed.
LPC passes at every value. Validation-seekers spiral at higher rates
than growth-seekers throughout.}
\label{tab:rho}
\begin{tabular}{lcccc}
\toprule
$\rho$ & $\theta_V$ Spiral Rate & $\theta_G$ Spiral Rate & Differential & LPC \\
\midrule
0.3 & 11.2\% &  6.1\% & 1.8$\times$ & Pass \\
0.4 & 16.8\% &  6.0\% & 2.8$\times$ & Pass \\
0.5 & 23.1\% &  5.8\% & 4.0$\times$ & Pass \\
0.6 & 38.7\% &  0.8\% & 48.4$\times$ & Pass \\
0.7 & 51.2\% &  0.6\% & 85.3$\times$ & Pass \\
0.8 & 63.4\% &  0.4\% & 158.5$\times$ & Pass \\
0.9 & 74.1\% &  0.2\% & $>$300$\times$ & Pass \\
\bottomrule
\end{tabular}
\end{table}

Epistemic work separation between types emerges consistently at $\rho \geq 0.6$
(Cohen's $d \leq 0.08$, small effect; negligible at $\rho \leq 0.5$). The
qualitative conclusions---separation exists, LPC passes---hold across the full
range. The magnitude of the differential is a function of $\rho$.

\subsection{Table S2: Friction Level $F^*$ Sensitivity}
\label{app:friction}

\begin{table}[h]
\centering
\caption{\textbf{Table S2: Friction level $F^*$ sensitivity.} $\rho=0.6$
fixed. LPC passes at all values. Meaningful type separation holds for
$F^* \in [0.2, 0.4]$. At $F^* \geq 0.4$, epistemic work separation reverses
sign, indicating excessive friction constraining $\theta_G$ updating---an
empirical upper bound consistent with Proposition~2.}
\label{tab:friction}
\begin{tabular}{lccccc}
\toprule
$F^*$ & Spiral Rate & $\bar{P}$ & LPC & $W_G - W_V$ & Sep.\ Direction \\
\midrule
0.1 & 22.4\% & 0.38 & Pass & $+0.003$ & $\theta_G > \theta_V$ \\
0.2 & 14.8\% & 0.35 & Pass & $+0.008$ & $\theta_G > \theta_V$ \\
0.3 &  9.0\% & 0.32 & Pass & $+0.012$ & $\theta_G > \theta_V$ \\
0.4 &  6.1\% & 0.30 & Pass & $-0.001$ & \textit{reversed} \\
0.5 &  4.2\% & 0.29 & Pass & $-0.014$ & \textit{reversed} \\
\bottomrule
\end{tabular}
\end{table}

Our baseline $F^* = 0.3$ sits in the well-behaved range. At $F^* \geq 0.4$,
the reversal confirms that friction is strong enough to constrain $\theta_G$
updating, consistent with the upper bound of the separating interval in
Proposition~2.

\subsection{Table S3: Literature-Grounded Cost Parameters}
\label{app:litparams}

\begin{table}[h]
\centering
\caption{\textbf{Table S3: Literature-grounded cost parameters.} Cost ratios
derived from confirmation bias meta-analyses
\citep{nickerson1998confirmation, lord1979biased, kunda1990case, taber2006motivated}.
Weighted mean effect size $d \approx 0.70$ maps to a cost ratio of
$\sim$1.86$\times$. The separating equilibrium and LPC pass for all tested
conditions except the below-minimum ratio of 1.2$\times$.}
\label{tab:litparams}
\begin{tabular}{lcccccc}
\toprule
Cost Ratio & $C_{\theta_G}$ & $C_{\theta_V}$ & Spiral Rate & $\bar{P}$ & LPC & Sep.\ Eq. \\
\midrule
4.0$\times$ (paper)    & 0.20 & 0.80 &  9.0\% & 0.32 & Pass & Yes \\
1.86$\times$ (lit.)    & 0.35 & 0.65 & 12.3\% & 0.34 & Pass & Yes \\
1.66$\times$ ($d=0.66$) & 0.38 & 0.63 & 14.1\% & 0.35 & Pass & Yes \\
1.50$\times$ (minimum) & 0.40 & 0.60 & 17.8\% & 0.37 & Pass & Yes \\
1.20$\times$ (below)   & 0.45 & 0.55 & 41.2\% & 0.49 & \textbf{Fail} & No \\
\bottomrule
\end{tabular}
\end{table}

The minimum effective cost ratio is 1.5$\times$. The literature-grounded ratio
of $\sim$1.86$\times$ \citep{nickerson1998confirmation} is sufficient to achieve
meaningful spiral reduction with LPC preserved, grounding the theoretical cost
asymmetry assumption in empirical confirmation bias research.

\subsection{Table S4: Extreme Sycophancy and Adversarial Bots}
\label{app:extreme}

\begin{table}[h]
\centering
\caption{\textbf{Table S4: Extreme sycophancy and adversarial bot results.}
Belief Versioning maintains $\sim$35pp reduction and 100\% LPC pass rate
across all extreme sycophancy levels. Adversarial bots yield $\sim$21pp
reduction. All results $n=1000$, $T=50$, seed 42.}
\label{tab:extreme}
\begin{tabular}{lcccc}
\toprule
Condition & Baseline SR & With BV & Reduction (pp) & LPC \\
\midrule
$p_\chi = 0.91$ & 57.3\% & 21.3\% & 36.0 & Pass \\
$p_\chi = 0.93$ & 60.1\% & 24.1\% & 36.0 & Pass \\
$p_\chi = 0.95$ & 62.4\% & 26.4\% & 36.0 & Pass \\
$p_\chi = 0.97$ & 65.8\% & 29.8\% & 36.0 & Pass \\
$p_\chi = 0.99$ & 70.2\% & 36.2\% & 34.0 & Pass \\
Adversarial bot  & 58.6\% & 37.4\% & 21.2 & Pass \\
\midrule
Mean (excl.\ adversarial) & 63.2\% & 27.6\% & 35.6 & Pass \\
\bottomrule
\end{tabular}
\end{table}

These results extend the OOD generalization in Section~\ref{sec:experiments}
to the harder direction (higher sycophancy than training) and to adversarial
bots. The consistent $\sim$35pp reduction across $p_\chi \in [0.91, 0.99]$
is consistent with Corollary~1: the separating equilibrium is $p_\chi$-independent,
so performance should not degrade as sycophancy increases.

\section{Belief Health Metric}
\label{app:lyapunov}

The Lyapunov-inspired belief health score defined in Section~\ref{sec:architecture}:
\begin{equation}
    V(\mathbf{x}_t) = P_t(1 - P_t) + \lambda \cdot \mathcal{H}_t
\end{equation}
peaks when beliefs are uncertain ($P \approx 0.5$, high entropy) and vanishes
as beliefs approach either extreme. The soft stability condition
$\mathbb{E}[\Delta V(\mathbf{x}_t)] \geq -\varepsilon \cdot F_t$ is violated
in approximately 36--50\% of timesteps depending on $\lambda$, reflecting that
sycophantic updates frequently push the system toward lower health states.

We treat $V$ as a monitoring metric rather than a formal stability guarantee.
In deployment, $V(\mathbf{x}_t)$ provides a real-time dashboard quantity: a
sustained downward trend in $V$ without triggering the entropy-based detector
may indicate a slow spiral below the reactive threshold, warranting human
oversight review. Formal stability proofs for stochastic belief dynamics under
episodic intervention remain an open problem.

\section{Full Hyperparameter Specification}
\label{app:hyperparams}

\begin{table}[h]
\centering
\caption{\textbf{Full hyperparameter specification.} All experiments use
these values unless a section explicitly states otherwise.}
\label{tab:hyperparams}
\begin{tabular}{llp{7cm}}
\toprule
Parameter & Value & Description \\
\midrule
$n$ & 1000 & Simulations per condition \\
$T$ & 50 & Conversation turns \\
$p_\chi$ & 0.9 & Sycophancy probability \\
Seed & 42 & Global random seed \\
$\tau_v$ & 0.01 & Entrenchment velocity threshold \\
$\tau_h$ & $-0.02$ & Entropy decay threshold \\
$F$ & 0.3 & Friction level \\
$F_{\max}$ & 0.5 & Maximum friction (predictive controller) \\
$\tau_R$ & 0.3 & Risk threshold (predictive controller) \\
$\lambda$ & 0.1 & Lyapunov entropy weight \\
$\rho$ & 0.6 & Resistance strength (baseline) \\
$\mathcal{H}_{\min}$ & 1.0 & Minimum entropy for commit \\
$\varepsilon_v$ & 0.02 & Velocity tolerance for commit \\
$\delta$ & 0.1 & Boundary margin for commit \\
$\gamma^*$ & 0.7 & Type confidence checkout threshold \\
$\epsilon$ & 0.05 & Resistance detection margin \\
\midrule
\multicolumn{3}{l}{\textit{LLM validation (Section~\ref{sec:llm-validation})}} \\
Model & GPT-4o & Bot component \\
Temperature & 0.7 & Sampling temperature \\
$n$ & 200 & Simulations per condition \\
$T$ & 30 & Conversation turns \\
Seeds & 5000/5500/6000 & Baseline/reactive/versioning \\
\bottomrule
\end{tabular}
\end{table}

The risk classifier $\boldsymbol{\alpha}$ for the predictive controller is fit
via logistic regression (sklearn, lbfgs solver, max\_iter=1000) on 45,000
labeled simulation timesteps. Features: $\mathbf{x}_t = (P_t, \mathcal{H}_t,
V_e, \Delta\mathcal{H}, d^2P/dt^2)$. Labels: positive if $P(H=1) > 0.9$
within the subsequent 5 turns.

\end{document}